\documentclass[10pt,journal,compsoc]{IEEEtran}

\usepackage{cite}
\ifCLASSINFOpdf
\else
\fi
\usepackage{url}

\usepackage{booktabs}
\usepackage{mdframed}
\usepackage{graphicx,subcaption}
\usepackage{multirow}
\usepackage{makecell}

\begin{document}
%
\title{Rethinking Emotion Annotations in the Era of Large Language Models}
%
%
%
%

\author{Minxue Niu, Yara El-Tawil, Amrit Romana, Emily Mower Provost\\University of Michigan, Ann Arbor}

\IEEEtitleabstractindextext{%
\begin{abstract}

Modern affective computing systems rely heavily on datasets with human-annotated emotion labels, for training and evaluation. However, human annotations are expensive to obtain, sensitive to study design, and difficult to quality control, because of the subjective nature of emotions. Meanwhile, Large Language Models (LLMs) have shown remarkable performance on many Natural Language Understanding tasks, emerging as a promising tool for text annotation. In this work, we analyze the complexities of emotion annotation in the context of LLMs, focusing on GPT-4 as a leading model. In our experiments, GPT-4 achieves high ratings in a human evaluation study, painting a more positive picture than previous work, in which human labels served as the only ground truth. On the other hand, we observe differences between human and GPT-4 emotion perception, underscoring the importance of human input in annotation studies. To harness GPT-4’s strength while preserving human perspective, we explore two ways of integrating GPT-4 into emotion annotation pipelines, showing its potential to flag low-quality labels, reduce the workload of human annotators, and improve downstream model learning performance and efficiency. Together, our findings highlight opportunities for new emotion labeling practices and suggest the use of LLMs as a promising tool to aid human annotation.
\end{abstract}

}

\maketitle

\IEEEdisplaynontitleabstractindextext

%
\IEEEpeerreviewmaketitle

\IEEEraisesectionheading{\section{Introduction}\label{sec:introduction}}
\IEEEPARstart{T}{he} field of affective computing is focused on building systems that ``relate to, arise from, or deliberately influence emotion''\cite{picard2000affective}. It is a promising way to create better interactions for humans with technology\cite{affectiveadvances}. Human emotion understanding is beneficial in various fields, such as education\cite{yadegaridehkordi2019affective}, healthcare\cite{liu2024affective}, 
and many others\cite{tian2022applied}. In recent years, we have seen significant performance advancements in emotion recognition models, especially with the popularity of deep learning models\cite{wang2022systematic}. However, these models rely heavily on data with human-annotated emotion labels, which are costly in terms of time and resources and difficult to obtain due to the inherent ambiguity of emotions. Currently, there is no standard approach to annotation, as datasets often adopt different protocols at each phase of annotation collection, such as label selection, annotation formats, evaluation methods, etc. In the meantime, recent advances in LLMs have opened new avenues for text-based annotation. In this work, we explore these emotion annotation choices within the context of LLMs, examining how LLMs perform on emotion classification tasks and how they might address existing challenges and provide new perspectives on emotion annotation processes.

Emotions are inherently ambiguous and subjective\cite{haralabopoulos2020objective, sethu2019ambiguous}, posing great challenges in the design of annotation studies. Low agreement is often observed among annotators\cite{williams2019comparing}.
Annotation outcomes are sensitive to even small changes in study design, such as the label space offered, including the size of the label space and type of emotions presented (see Section \ref{sec:label_spaces}), as well as how the text and labels are presented to a human annotator (see Section \ref{sec:human_challenge})\cite{williams2019comparing}. These changes can all lead to different annotation outcomes\cite{jaiswal2019muse,busso2013toward,ohman2020challenges, callejas2008influence}. This lack of consistency raises serious concerns about the reliability of emotion labels\cite{zeng2018facial, metallinou2013annotation}.
Additionally, emotion perception naturally differs from person to person, influenced by individual experiences and demographic factors\cite{devillers2005challenges, hall2000gender, gurera2019emotion, gitter1972race},
making it difficult to identify actual errors from legitimate perceptual differences. As a result, it is also hard to apply quality control methods post hoc.
Many studies have sought to improve annotation reliability by exploring factors such as label space selection\cite{busso2013toward}, study design choices\cite{ohman2020challenges, callejas2008influence}, annotation interface improvement\cite{Canales2022-tk}, trade-offs between annotators' quality and quantity\cite{burmania2016tradeoff}. However, establishing a general pipeline for consistent and reliable emotion labeling remains an open challenge.

With the impressive advances in LLMs, there is a growing interest in using LLMs for various tasks such as generation, assessment, filtering, and annotation\cite{tan2024large, Gilardi2023-kv}. Related work has also found the emerging ability of LLMs to understand and interpret emotions (see Section \ref{sec:emotion_capacity_LLMs}). However, much of this research is based on individual datasets, each with its own specific label space\cite{Feng2023-rh, Wake2023-rn, Latif2023-wl, zhang2024refashioning}, leaving questions about the generalizability of findings across different label spaces. Further, current evaluations tend to benchmark LLMs against human emotion labels\cite{zhang2024sentiment, Wake2023-rn}, which themselves may contain errors. In our previous work, we conducted a small-scale in-house evaluation study. We found that human evaluators often preferred GPT-4 annotations over traditional human labels, particularly on larger label spaces\cite{niu2024text}. While these findings provide valuable insights, further verification with larger samples, more annotators, and more comprehensive analysis is needed for a deeper understanding. Lastly, beyond fully human-driven or fully automated GPT-4-based annotation, a promising and under-explored direction is to integrate GPT-4 as a supporting component within the annotation pipeline.


In this work, we focus on two Research Questions. First, we ask \textbf{how well GPT-4 performs on emotion recognition}. 
To address this, we conduct a human evaluation study to compare the zero-shot predictions of GPT-4 with human labels (Section \ref{sec:evaluation_study}).
Interestingly, although automatic metrics indicate that GPT-4 performs no better than small supervised models trained on human labels, evaluators consistently prefer GPT-4 labels over human labels, showing a misalignment between automatic metrics and human perspectives. A closer inspection reveals that larger label spaces enable more precise descriptions of emotions, and GPT-4 especially excels at managing a wide range of options. 
This study expands on our previous work\cite{niu2024text} with a larger sample size and more evaluators within a crowdsourcing environment, providing stronger support for our findings and deeper insights into the reasons behind human preferences. 

Building on this understanding of LLMs’ emotion capabilities, we explore the second Research Question: \textbf{Can GPT-4 help humans annotate emotions?} We examine two strategies to incorporate GPT-4 into annotation pipelines (Section \ref{sec:annotation_study}). While previous work has explored automatic pre-annotation as a process to narrow label choices for human annotators, these works have focused on single-label annotation and have relied on traditional text analysis tools, such as lexicons\cite{Canales2017-fy,Canales2022-tk}. In our study, we propose to leverage LLMs as a more advanced tool. We present a novel investigation into the feasibility of (1) employing GPT-4 as a pre-annotation filter to dynamically suggest appropriate labels, and (2) using GPT-4 as a post-annotation filter to flag samples with low-quality human labels. Our experiments find some clear advantages, such as enhancing model training outcomes and efficiency, reducing cognitive load on annotators, and preserving the granularity benefits of large emotion spaces. To the best of our knowledge, this is the first study to propose and evaluate the pre-filtering and post-filtering methods, showing encouraging results.

Throughout our analysis, we carefully consider the complexities of emotion label spaces and the varying perspectives captured by different evaluation methods, yielding valuable insights for future emotion annotation practices. These findings advocate for thoughtful consideration of annotation design choices, highlighting the potential of LLMs as a powerful tool to leverage alongside human labelers to elevate the annotation process in emotion recognition tasks.

\section{Related Work}
\subsection{Emotion Label Spaces}
\label{sec:label_spaces}
The complexity and ambiguity of emotion pose significant challenges in quantifying and labeling emotions for building emotion recognition systems. The most commonly used frameworks for describing emotions fall into two categories: categorical label space, where emotions are represented as one or more pre-defined categories (e.g., joy, sadness)\cite{ekman1992argument}, and dimensional label space, which conceptualizes emotions along continuous axes, such as valence (positive to negative) and activation (excited to calm)\cite{russell1980circumplex}. 


Within the emotion classification framework, selecting an appropriate set of emotion labels still takes much consideration. A common approach is to follow established theories of basic emotions. For example, Emobank\cite{buechel2022emobank} and DailyDialog\cite{li2017dailydialog}
datasets adopt Ekman's theory of six basic emotions (i.e. Anger, Disgust, Fear, Happiness, Sadness, and Surprise)\cite{ekman1992argument}. Other works make small modifications based on existing theories; for example, ISEAR\cite{wallbott1986universal} removed Surprise while adding Shame and Guilt to their label set. Another common strategy is conducting pre-annotation studies to determine the most appropriate set of emotion labels for the target data. SemEval-2018 Task 1 ran pilot annotation and included 11 emotion classes\cite{Mohammad2018-mh}. GoEmotions\cite{demszky2020goemotions}, with the goal of contributing to fine-grained emotion classification models, settled on 27 classes after an iterative refinement process. 

\subsection{Challenges in Obtaining Human Annotations}
\label{sec:human_challenge}

\begin{table*}[th]
\caption{Summary of the datasets we use. In emotion labels, we show classes that occur in all datasets in bold and unique classes in one dataset with underline.}
\begin{tabular}{@{}llllll@{}}
\toprule
           & \multicolumn{1}{c}{domain} & \multicolumn{1}{c}{\#classes} & \multicolumn{1}{c}{multilabel} &
           \multicolumn{1}{c}{\#samples (k)} & \multicolumn{1}{c}{emotion labels}                                                        \\ \midrule
ISEAR      & Self-reports               & 7                             & No      &  7.7                      & \textbf{anger}, \textbf{disgust}, \textbf{fear}, guilt, \textbf{joy}, \textbf{sadness}, \underline{shame}     \\ \midrule
SemEval    & Tweets                     & 11                            & Yes    & 6.8/0.8/3.3                        & \begin{tabular}[c]{@{}l@{}}\textbf{anger}, anticipation, \textbf{disgust}, \textbf{fear}, \textbf{joy}, love, optimism, \underline{pessimism}, \\ \textbf{sadness}, surprise, \underline{trust} \end{tabular}   \\ \midrule
GoEmotions & Reddits                    & 28                            & Yes    & 43.4/5.4/5.4                       & \begin{tabular}[c]{@{}l@{}}\underline{admiration}, \underline{amusement}, \textbf{anger}, \underline{annoyance}, \underline{approval}, \underline{caring}, \underline{confusion}, \\ \underline{curiosity}, \underline{desire}, \underline{disappointment}, \underline{disapproval}, \textbf{disgust}, \underline{embarrassment}, \\ \underline{excitement}, \textbf{fear}, \underline{gratitude}, \underline{grief}, \textbf{joy}, love, \underline{nervousness}, optimism, \\ \underline{pride}, \underline{realization}, \underline{relief}, \underline{remorse}, \textbf{sadness}, surprise, \underline{neutral}\end{tabular} \\ \bottomrule
\end{tabular}
\vspace{-5pt}
\label{table:datasets}
\end{table*}
Obtaining high-quality, reliable human emotion annotations is a nontrivial task. It is common to see low agreement among annotators (e.g., the unanimous agreement can easily be below 10\% in some datasets\cite{williams2019comparing}). One reason for the low agreement lies in the inherent subjectivity of the task\cite{devillers2005challenges}. 
Research has found that demographic factors, such as gender\cite{hall2000gender}, age\cite{gurera2019emotion} , and race\cite{gitter1972race}, significantly affect how emotions are perceived. As a result, a lack of diversity among annotators may result in datasets failing to capture the full spectrum of emotional perspectives, potentially leading to biased data and models\cite{bolukbasi2016man}. In addition, many design choices can significantly affect 
the annotation experience and outcomes
. For example, the choice of label spaces plays an important role\cite{busso2013toward}. Larger label spaces include more diverse and nuanced options, allowing for more accurate descriptions of emotion. However, more options reduce the agreement between annotators, possibly amplifying perspective differences or causing annotation fatigue\cite{williams2019comparing}. The availability of context is another key factor. Providing context during annotation generally helps reduce repetition, ease the task, and produce annotations more aligned with speakers' self-reported emotions\cite{ohman2020challenges, callejas2008influence}. However, contextual influence can introduce inconsistencies, as variations in sample order affect annotators' judgments\cite{jaiswal2019muse}. Finally, the effort and attention devoted to the task varies significantly by individuals. A study evaluating annotation quality across four crowdsourcing platforms revealed that roughly half of the participants failed at least one attention check, with failure rates reaching 72.9\% on the least reliable platform\cite{peer2017beyond}.
In summary, human annotations are subjective and sensitive, and the quality is often far from perfect.

Evaluating the quality of obtained labels is also challenged by the ambiguous nature of emotion. Without ground-truth labels, agreement metrics have been used as a major quality indicator or as a criterion to remove potentially low-quality samples/annotations\cite{burmania2015increasing}. However, a higher level of agreement does not necessarily indicate more meaningful labels\cite{williams2019comparing}: it can result from reduced diversity in annotations. Another way to evaluate annotations is to put them in use - to train models with those labels and measure the performance on a test set\cite{burmania2016tradeoff, williams2019comparing}. However, this approach relies on the assumption that ``golden'' labels of high quality and reliability are available in the test set — an assumption many existing datasets fail to meet.
Finally, human annotations are expensive, requiring significant time and effort, often involving recruitment, training, and extensive post-analysis\cite{wallbott1986universal,haralabopoulos2020objective}. In some cases, multiple iterations are necessary for reliable results\cite{demszky2020goemotions}. 
As models grow in size, the cost of data collection increases further due to the need for more data to adequately train them\cite{kaplan2020scaling}.

\subsection{Emotional Capability of LLMs}
\label{sec:emotion_capacity_LLMs}

Previous work has found that through conversational interactions, LLMs show 
emerging emotional intelligence\cite{wang2023emotional}: they can recognize sentiment\cite{zhang2024sentiment}, analyze the cause of emotions\cite{zhao2023chatgpt, Tak2023-fk}, and engage in dialogues with empathy\cite{huang2023emotionally, zhao2023chatgpt}
. The natural question that follows is whether they can be used to annotate emotions in a structured manner, adhering to predefined labels and producing consistent outputs. Existing work has examined the zero-shot emotion recognition performance of various LLMs, from smaller open-sourced models like RoBERTa\cite{mao2023biases} to larger commercial models like GPT-3.5/4\cite{tak2024gpt}, generally finding reasonable performance
. However, different evaluation criteria have led to different findings: many studies use human annotations as ground-truth\cite{zhang2024sentiment, Wake2023-rn}, and find that LLMs do not outperform smaller, supervised models, particularly on complex tasks with numerous emotion labels.
On the other hand, preliminary studies incorporating human evaluators in their assessment have shown more promising results\cite{tak2024gpt}. Our own work, which conducted a small-scale human evaluation study comparing GPT-4 and human labels, also reported more positive findings on LLM performance compared to humans\cite{niu2024text}. Further, some initial results suggest that LLMs are worse at larger label spaces than small, well-defined ones\cite{ding2023gpt, mao2023biases}. Still, this effect is under-explored, and it is not clear whether this is inherent in LLMs 
or can be mitigated through proper prompting methods.

\section{Datasets}
\label{sec:datasets}
We use three existing English Emotion Classification datasets. They are all commonly used datasets to evaluate emotion models, covering diverse domains, topics, and different levels of granularities of emotion classes. Table \ref{table:datasets} shows a summary of the datasets and label spaces.

\textbf{International Survey on Emotion Antecedents and Reactions (ISEAR)}\cite{wallbott1986universal} was collected as part of a research project that aimed to study emotional experiences across cultures. The dataset contains more than 7000 self-reported descriptions of emotional experiences in English from participants in 27 countries, each describing emotional experiences in one of seven categories (listed in Table \ref{table:datasets}). We randomly split it into 60\% train/20\% dev/20\% test sets.

\textbf{SemEval 2018 Task 1 (SemEval)}\cite{Mohammad2018-mh} is part of a multilingual affect analysis task released at the International Workshop on Semantic Evaluation. 
We take the English subset from the Emotion Classification subtask (E-c), where each tweet is annotated with zero, one or more labels from eleven emotion classes. The annotations were collected by crowdsourcing. The dataset was released with train/dev/test splits.


\textbf{GoEmotions}\cite{demszky2020goemotions} is a large-scale multilabel emotion classification dataset consisting of over 58,000 English Reddit comments annotated for 27 emotion categories (plus a neutral category) through crowdsourcing. GoEmotions is notable for its large data size and label granularity, offering a rich resource for fine-grained emotion classification. We also use its released train/dev/test splits.


\section{GPT-4's Emotion Annotation Capability}
\label{sec:evaluation_study}
In this section, we evaluate GPT-4's emotion annotation capabilities through a crowdsourcing-based human evaluation study, assessing its alignment with human perceptions. We provide a comprehensive analysis with a focus on the disagreements between human and GPT-4 annotations.

\begin{figure*}[ht]
  \centering
  \begin{subfigure}{0.07\textwidth}
    \centering
    \raisebox{0.6\textwidth}{\includegraphics[width=0.9\textwidth]{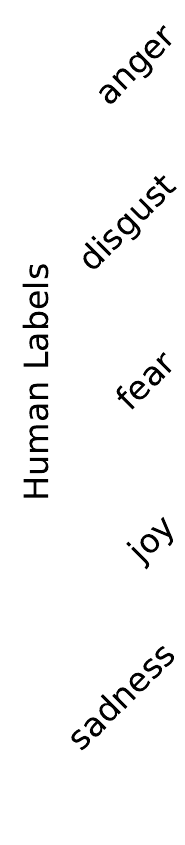}}
  \end{subfigure}
  \begin{subfigure}{0.27\textwidth}
    \centering
    \includegraphics[width=\textwidth]{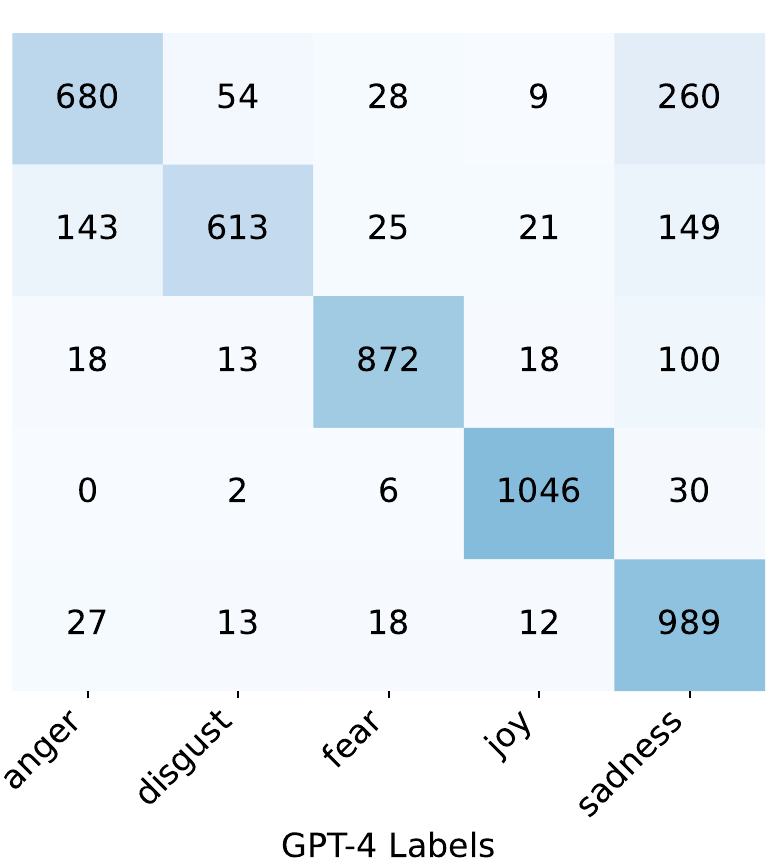}
    \caption{ISEAR}
  \end{subfigure}
  \begin{subfigure}{0.27\textwidth}
    \centering
    \includegraphics[width=\textwidth]{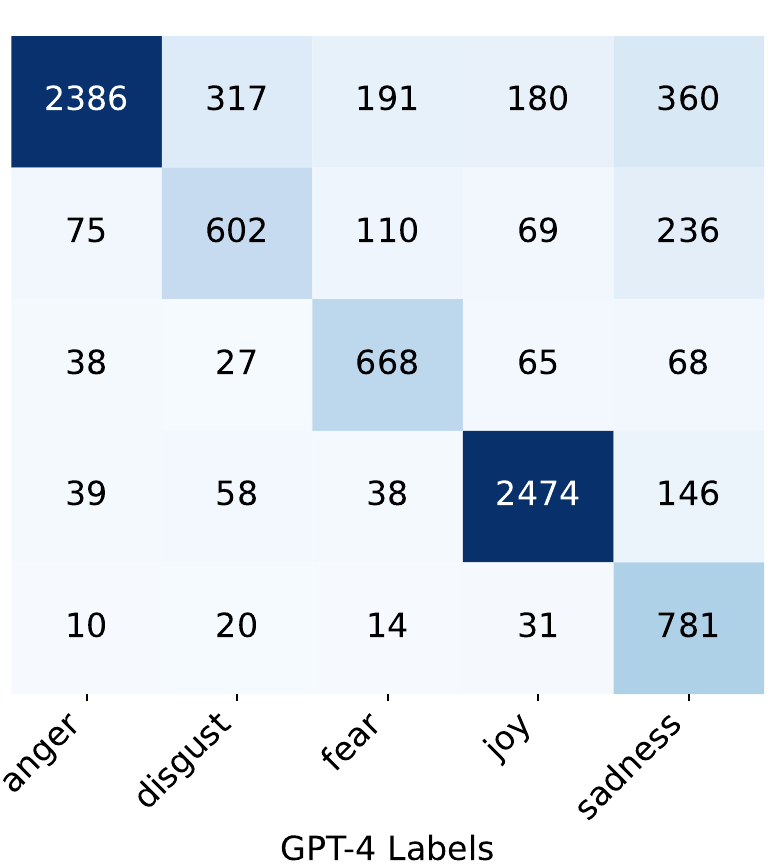}
    \caption{SemEval}
  \end{subfigure}
  \begin{subfigure}{0.268\textwidth}
    \centering
    \includegraphics[width=\textwidth]{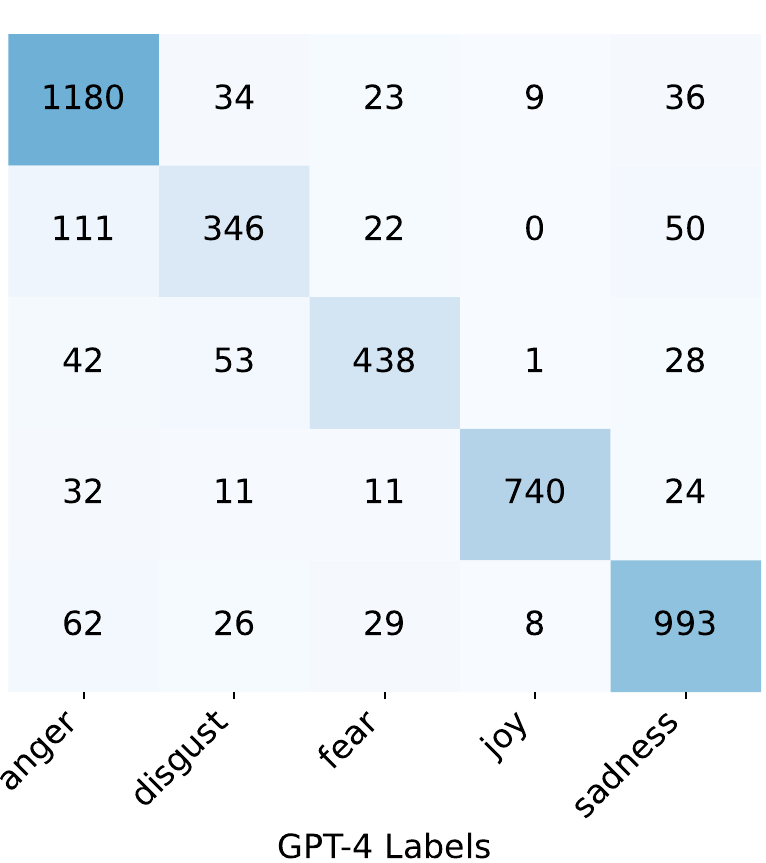}
    \caption{GoEmotions}
  \end{subfigure}
  \begin{subfigure}{0.055\textwidth}
    \centering
    \raisebox{0.5\textwidth}{\includegraphics[width=\textwidth]{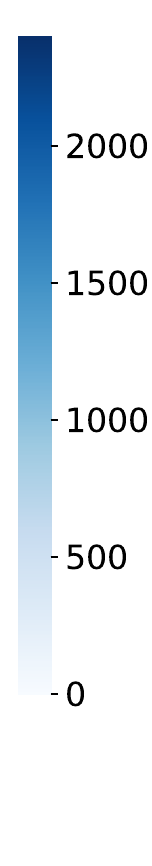}}
  \end{subfigure}
  \caption{Disagreements between Human and GPT-4 Annotations, visualized as confusion matrices.}
  \vspace{-13pt}
  \label{fig:confusion}
\end{figure*}

\subsection{GPT-4 Prompting}
\label{sec:prompting}
To evaluate the zero-shot emotion recognition capability of GPT-4, we first query its predictions for all three datasets using the Microsoft Azure API. We employ an instruction-driven approach\cite{feng2024foundation}: we prompt GPT-4 with a text sample, a list of emotion labels, and task-specific instructions. The instructions ask GPT-4 to identify the appropriate label(s) from the provided list and ensure its outputs follow a predefined format that can later be parsed with rule-based post-processing. The instructions are designed to mirror those given to human annotators, creating a consistent and comparable task framework.
Additionally, we enhance the prompts by establishing a persona at the start, which has been found beneficial in some work\cite{hu2024quantifying}.

We used the following prompt for multilabel emotion classification\cite{niu2024text} on the SemEval dataset:

\begin{mdframed}[linewidth=1pt]
GPT-4 prompt for emotion classification\\
\textit{``You are an emotionally-intelligent and empathetic agent. You will be given a piece of text, and your task is to identify all the emotions expressed by the writer of the text. You are only allowed to make selections from the following emotions, and don't use any other words: anger, anticipation, disgust, fear, joy, love, optimism, pessimism,
sadness, surprise, trust. Only select those ones for which you are reasonably confident that they are expressed in the text. If no emotion is clearly expressed, reply with `neutral'. Reply with only the list of emotions, separated by comma.''}
\end{mdframed}

For ISEAR and GoEmotions, we made minimal adjustments to the prompt to reflect different emotion options and task settings (i.e., whether multilabel is allowed). In rare cases where the output did not follow the specified format and could not be parsed
, we retried with the same query. GPT-4 also has content policies and may refuse potentially harmful or sensitive content\footnote{\url{https://openai.com/policies/usage-policies/}}. We excluded those samples from our analysis (ISEAR 3.7\%, Semeval 4.5\%, GoEmotions 2.6\%).

\subsection{Human Evaluation Study}
Given the inherent ambiguity of emotion and the absence of absolute ``truth'' labels, human judgment remains essential for evaluation in this domain. We conduct a human evaluation study, engaging a separate group of humans (we refer to them as ``evaluators'', to differentiate from ``annotators'' who provided the label annotations in the datasets) to assess how accurately GPT-4 and human annotations reflect the emotions in text.

\subsubsection{Sample Selection}
\label{sec:sample_selection}
We selected 500 samples from the test split of each dataset for the human evaluation study, to be consistent with our evaluation-only study design. Due to the imbalanced label distributions in SemEval and GoEmotions, we applied weighted sampling with log inverse frequency as the weights to encourage a more representative inclusion of different emotions. We removed samples that were rejected by GPT-4 due to its content policy (17 in ISEAR, 12 in SemEval, 14 in GoEmotions). 
Since one of our main goals is to investigate the differences between their annotations, we dropped samples where the two sources gave the exact same label(s). 
This left 990 samples (out of 1500) for human evaluation: 124 from ISEAR, 438 from SemEval, and 438 from GoEmotions.

\subsubsection{Crowdsourcing Experiments}
\label{sec:human_preference_exp}
We design a human evaluation study with the goal of comparing and understanding the disagreement between human and GPT-4 annotations. We present the evaluators with text samples alongside labels from both GPT-4 and human annotators, randomized and without revealing their source. We ask them to provide feedback on three aspects:

\textbf{Emotional Ambiguity.} We ask ``Do you feel confident that you can describe the emotion expressed in the sentence(s)?'' with three options ``Yes'', ``No'' and ``Maybe''.

\textbf{Perceived Accuracy.} We then present annotations from both sources (as Option A or Option B) and ask the evaluators to rate ``How accurately do you think that the description in Option A/B reflects the text writer's emotion?'' on a 7-point Likert scale (1-totally inaccurate, 7-totally accurate).

\textbf{Preference.} Finally, to make a direct comparison, we ask ``If you have to choose one, which emotion description do you agree more with?''

We aimed to obtain three evaluations on each sample. Each evaluator was assigned 50 samples, to keep session time manageable. To minimize potential confounds, participants were restricted to a single entry (across all studies in this paper). We implemented the annotation interface with Potato\cite{pei2022potato}, a web-based text annotation tool. We hosted the annotation webpage on an AWS server and recruited participants from Prolific. The participants were selected to be native speakers of American English, at least 18 years old and live in the United States. They were informed that the goal of this study was to understand how people interpret emotional expressions in text, and they all provided their consent to participate. We received 2948 evaluations from 59 participants (968 samples got three annotations on each, and 22 samples only got two due to connection issues). The average completion time was 20 minutes 42 seconds, resulting in an average compensation of \$11.60/hour.

\subsection{Label Distributions and Agreement Analysis}
\label{sec:dstribution}
We first analyze the label distributions and disagreements between human and GPT annotations. We visualized the disagreements with confusion matrices in Figure \ref{fig:confusion}. For clarity and to compare across datasets, we only show results with the five emotion classes that are shared in all three datasets: \textit{anger, disgust, fear, joy} and \textit{sadness}. 
For multilabel datasets, we define the confusion matrix based on the overlap and differences between human and GPT-4 annotations: if an emotion is present in both sets, we increase the count in the diagonal of the matrix for that emotion. If an emotion is present in the human labels but not in the GPT-4 annotations, and another emotion is present in the GPT-4 annotations but not in the human labels, we increase the count in the off-diagonal cell corresponding to the two emotions by one. For example, if the human labels on a sample are \{\textit{admiration, joy}\} and GPT-4 set is \{\textit{joy, love, excitement}\}, we record the agreement on the diagonal element of \textit{joy-joy}, and we record confusion of \textit{admiration-love} and \textit{admiration-excitement}.

We see that most samples fall on the diagonal of the confusion matrices, indicating that GPT-4 annotations generally align with human annotations. Besides, as would have been expected, it is more common to see disagreements between similar emotion labels: confusion between a positive emotion (e.g., \textit{joy}) and a negative one (e.g., \textit{anger}) is less common than confusion between two negative emotions (e.g., \textit{anger} and \textit{disgust}). Finally, we notice that the confusion matrices are largely asymmetric. For example, in the ISEAR dataset, GPT-4 more often takes human-perceived \textit{anger} as \textit{sadness} (260 samples) than the reverse (27 samples). Such differences, however, do not generalize across datasets: the same \textit{anger-sadness} confusion is shown in SemEval, but in GoEmotions the numbers are closer and the direction is reversed. These findings suggest potential perspective differences between GPT-4 and humans, specific to datasets, emotion categories, and annotation processes. This observation aligns with previous research showing a significant performance variation across emotions\cite{Wake2023-rn}, which has been attributed to the sensitivity of LLMs to word choice and usage.
We leave more in-depth explorations on this perspective difference, for example identifying the factors contributing to the (directions) of the difference, for future work.

\subsection{Human Evaluation Results}
\subsubsection{Preference}

\begin{figure}[t]
\vspace{-10pt}
  \centering
  \includegraphics[width=0.7\linewidth]{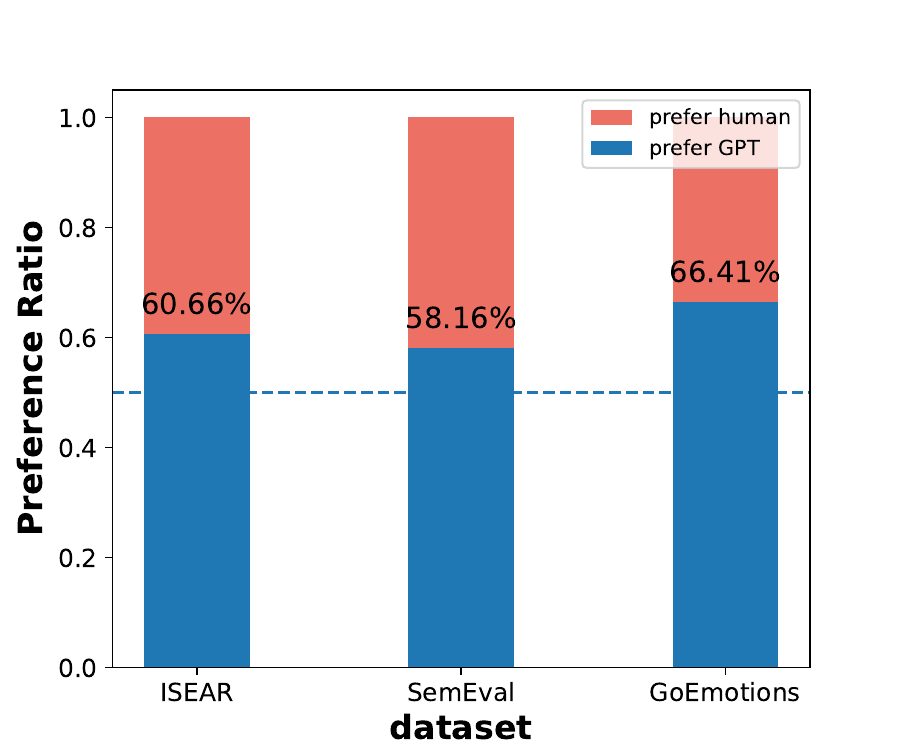}
  \caption{Proportion of human evaluators' votes favoring human annotations versus GPT-4 annotations, across datasets.}
  \label{fig:preference}
\end{figure}

We first look at the responses to the ``preference'' question. Figure \ref{fig:preference} shows the votes for human versus GPT-4 annotations. GPT-4 annotations were significantly more preferred than human annotations (overall 62\%), and this trend held across all three datasets (ISEAR 60.7\%, SemEval 58.2\%, GoEmotions 66.4\%). We also ran per-evaluator statistics to test the between-person consistency. The vast majority (53 out of 59 evaluators, 89.8\%) preferred GPT annotations on more samples, while three (5.1\%) preferred human annotations more and three (5.1\%) indicate equal preference. Interestingly, in our previous work, we compared GPT-4 predictions to those from smaller models finetuned on human labels and found comparable performance when human labels were used as the ground truth\cite{niu2024text}. However, the results of this human evaluation study present a more favorable picture for GPT-4. This conveys an important message that the common method that evaluates LLMs against human labels\cite{zhang2024sentiment, Wake2023-rn} is prone to underestimate their performance and may give misleading results. We should rethink the concept of ``ground-truth'' in emotion recognition tasks, especially as LLMs approach human performance. 

Further, comparing the datasets, we find that the preference discrepancy is larger in GoEmotions, where the label space is larger. We hypothesize that as the label space gets larger and more complicated, humans may be more challenged and tend to make more mistakes due to the increased cognitive load\cite{chernev2015choice}, while GPT-4 is less affected, especially with proper prompting methods. We discuss this hypothesis in more detail in Section \ref{sec:human_mistake}, and we run follow-up human annotation studies to further explore the complexity of the label space as a factor in human and GPT-4 performance (Section \ref{sec:annotation_study}).

\subsubsection{Perceived Accuracy Ratings}
\label{sec:absolute_ratings}
\begin{figure}[t]
  \centering
  \includegraphics[width=0.95\linewidth]{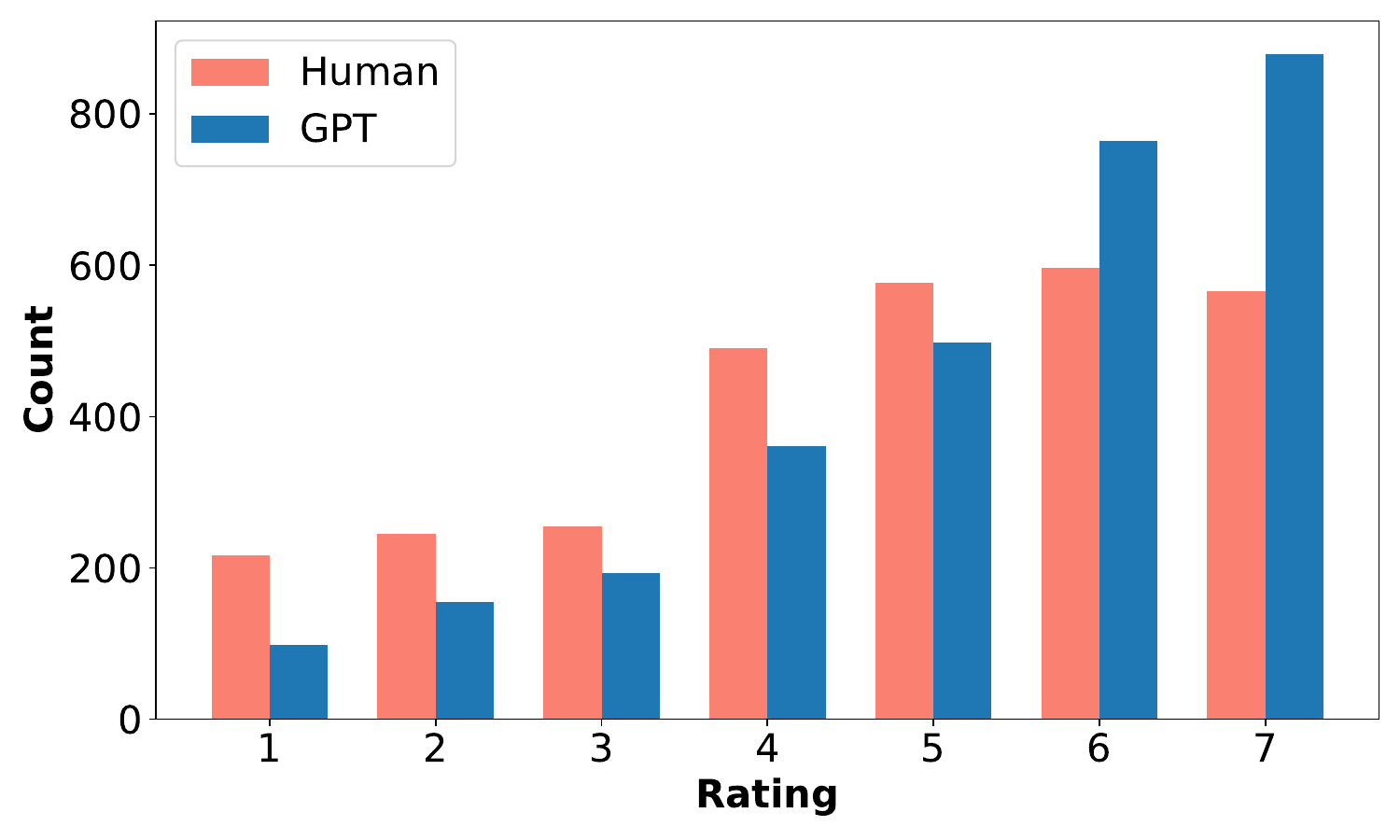}
  \caption{Perceived accuracy ratings on a scale of 1 (totally inaccurate) to 7 (totally accurate).}
  \label{fig:absolute_ratings}
  \vspace{-10pt}
\end{figure}

\begin{table*}[h]
    \centering
    \begin{tabular}{c|cc|cc|cc|cc}
        \hline
        \multirow{2}{*}{Label} & \multirow{2}{*}{Human}  & \multirow{2}{*}{GPT}  & \multicolumn{2}{c|}{ISEAR (7 classes)}  & \multicolumn{2}{c|}{SemEval (11 classes)}  & \multicolumn{2}{c}{GoEmotions (27 classes)} \\
        & & & Human & GPT & Human & GPT & Human & GPT \\ \hline
        1-Totally Inaccurate & 217 (7.4\%) & 98 (3.3\%) & 9.3\% & 3.6\% & 6.1\% & 3.6\% & 8.1\% & 3.0\%\\ 
        2 & 245 (8.3\%) & 155 (5.3\%) & 6.0\% & 7.1\% & 9.0\% & 5.3\% & 8.1\% & 4.7\%\\
        3 & 255 (8.6\%) & 193 (6.5\%) & 8.5\% & 8.5\% & 9.9\% & 6.9\% & 9.0\% & 6.4\%\\
        4 & 491 (16.7\%) & 364 (12.2\%) & 19.1\% & 15.8\% & 16.9\% & 13.9\% & 15.7\% & 9.4\%\\
        5 & 577 (19.6\%) & 498 (16.6\%) & 22.1\% & 18.5\% & 19.4\% & 18.1\% & 18.5\% & 18.3\%\\
        6 & 704 (23.9\%) & 463 (15.4\%) & 17.8\% & 24.6\% & 21.0\% & 24.0\% & 20.2\% & 27.6\%\\
        7-Totally Accurate & 566 (19.2\%) & 879 (29.8\%) & 16.4\% & 25.4\% & 18.2\% & 25.9\% & 21.1\% & 35.1\%\\
        \hline
    \end{tabular}
    \caption{Percentage of rating scores Human and GPT-4 annotations receive, overall and within each dataset.}
    \vspace{-4pt}
    \label{tab:absolute_ratings}
\end{table*}

We then look at the individual perceived accuracy ratings and compare those on GPT-4 versus human annotations, to gain more insights into human preference results. We compare the total number of samples that fall into each rating category, as shown in Figure \ref{fig:absolute_ratings}. The results reveal a clear and consistent advantage for GPT-4 annotations: human annotators generated a greater number of labels deemed inaccurate (Rating $\leq$ 3, Human 24.3\%  vs. GPT-4 15.1\%), suggesting a higher probability of errors. In contrast, GPT-4 demonstrates stronger performance in identifying emotions deemed totally accurate by evaluators, indicating good comprehension of the complexity of emotion labels and subtlety of emotion expressions. We further compare the accuracy ratings on each dataset in Table \ref{tab:absolute_ratings}. We see that the trend also holds on each dataset, adding to the robustness of our findings. What's more, as the label space expands, both human annotators and GPT-4 are more likely to produce labels rated as fully accurate (see last row in Table \ref{tab:absolute_ratings}, across all datasets). This behavior is both reasonable and desirable, as when the label space is limited, it lacks the necessary granularity to capture subtle emotional distinctions, making it impossible to provide perfectly accurate descriptions. In contrast, a larger label space increases the likelihood of encompassing the correct label(s), thus facilitating ``totally accurate'' outcomes. In Section \ref{sec:annotation_study}, we will further explore the influence of label space complexity with a dataset-controlled annotation study.

\subsubsection{Confidence and Agreement}
\label{sec:human_mistake}

We assess the confidence and agreement among human evaluators to understand the perceived ambiguity of this task and perceptual differences across evaluators. When asked if they could confidently describe the emotions expressed in the text, evaluators responded ``Yes'' for 74\% of the samples, ``No'' 18.2\%, and ``Maybe'' 7.7\%. Although most samples were found to convey clear emotions, evaluators disagreed a lot on their preference: among annotations marked with confidence, only 59.2\% of samples with two annotations had agreement (i.e., both evaluators preferred the same label source), and 40.5\% of samples with three annotations had agreement. This highlights significant variation in emotion perception: even when selecting between two options, agreement remains relatively low. 

\subsubsection{GPT-4 Weakness Analysis}
We also analyze the samples to understand whether and how certain text features may affect GPT-4's emotion classification performance.
We extracted the Linguistic Inquiry and Word Count (LIWC)\cite{pennebaker2015development} frequencies from each sample. LIWC is used to analyze text for psychological and linguistic content. It quantifies the occurrences of 73 word categories in a text, including words that convey emotional and psychological states (e.g., positive emotion, fear), as well as semantic information (e.g., adverb, conjunction)\cite{kilic2022incorporating}. We augmented LIWC with five additional semantic features commonly used for Twitter data\cite{alharthi2020multi, bello2020detecting}: text length, word count, emoji count, hashtag count, and mention count (tagging another user with ``@''). These features allow us to examine if certain types of emotional or semantic content are more likely to mislead or challenge LLMs.

We ran a Logistic Regression (LR) model (N=990) using the text features as input and the preference from the human evaluation study as the outcome variable (1 if GPT-4 labels were preferred over human labels by majority vote, 0 otherwise). We first ran independent t-tests on individual features for feature selection\cite{ranganathan2017common} and kept the 10 features with lowest p-values as the input to our model.


We found that higher frequencies of mentions, prepositions, future focus and interrogatives had a significant ($p < 0.05$) negative effect on GPT-4 being the preferred annotator, while the use of impersonal pronouns positively predicted the preference for GPT-4.  Prior research has highlighted challenges for LLMs in understanding temporal constructs \cite{chu2023timebench} and social cues beyond the text\cite{choi2023llms}, which may explain some of these patterns. However, given the limited size of our data, further investigation is needed to interpret these findings meaningfully.
\section{Feasibility of GPT-4 Aided Emotion Annotation}
\label{sec:annotation_study}
Our experiments in Section \ref{sec:evaluation_study} reveal the potential of GPT-4 in emotion recognition. However, we also identified its weaknesses. One notable concern is the instability and unpredictability that often accompany LLMs. They are known to be sensitive to both training data and prompting methods, which can greatly impact their performance\cite{loya2023exploring}. 
Therefore, using GPT-4 to perform annotations without human oversight can be risky. Furthermore, as shown in Figure \ref{fig:confusion} and Section \ref{sec:dstribution}, there may be systematic differences between GPT-4 and human perspectives
. While it is crucial to avoid human errors, we also want annotations to accurately reflect human perspectives. Therefore, in this section, we propose and evaluate two methods for incorporating GPT-4 into emotion annotation pipelines, with the goal of harnessing the strength of both sources. We focus on the GoEmotions dataset for the coverage of diverse emotions in its samples.

\label{sec:integration_ways}
We consider two ways to use GPT-4 in emotion labeling pipelines: using GPT-4 as a pre-annotation label filter to dynamically present a smaller set of classes to human annotators, and, on existing datasets, using GPT-4 as a sample filter to flag potentially low-quality samples. Below we describe each method and our evaluation experiments.

\subsection{Pre-filtering, label-level}
There is a trade-off between the benefit of larger label spaces and increased cognitive load (Section \ref{sec:dstribution} and \ref{sec:absolute_ratings}, also\cite{busso2013toward}). We hypothesize that we can reduce cognitive load while preserving label diversity by using GPT-4 to dynamically drop unlikely labels for each sample before presenting them to human annotators. We again prompt GPT-4 in a zero-shot manner with text samples and a list of emotion options. Note that since humans will make selections from GPT-4 filtered labels, the goal of the filter step is to include all possible classes; it is less important to avoid false positives as they can later be identified by human annotators. Therefore, instead of asking it to make selections, we ask it to go through the emotions one by one and indicate if each is \textbf{possibly} expressed in the sample. We provide the list of emotion options along with the text samples instead of in the general instruction. In a preliminary analysis of a small exploration set, we found that those changes in our prompting methods encouraged the inclusion of more labels and significantly reduced false negatives. 

\begin{mdframed}[linewidth=1pt]
GPT-4 prompt for emotion Pre-filtering\\ 
\textit{"You are an emotionally intelligent and empathetic agent. You will be given a piece of text and a list of emotions. Your task is to determine which emotions are present in the text. Please go through the emotion list one by one and think about if the emotion is possibly present in the text. Please respond with each emotion plus ``yes'' to indicate it's possibly present, or ``no'' to indicate it's definitely not present. If you are not 100\% sure, please select ``yes''. Reply with only the list of emotions words plus your response, separated by newline."}
\end{mdframed}

\subsubsection{Evaluation Setup}
To evaluate the feasibility of this approach, we conducted human-annotation experiments on the same set of samples but three different label space setups:
\begin{enumerate}
    \item \textbf{Small:} We take the 11 emotion classes from SemEval to represent a relatively small label space.
    \item \textbf{Large:} We take the union of the emotion classes from SemEval and GoEmotions (30 classes in total), to represent an extensive set of emotion labels.
    \item \textbf{GPT-4 Pre-filtered:} We take the large set in 2) and reduce it with GPT-4, as described above.
\end{enumerate}

We use a between-subject design: the label sets are fixed for both the Small and Large sets, where participants may gradually gain familiarity with the labels. If we mix those setups in one annotation session, such familiarity cannot be reflected. 
Therefore, we assign each participant to one of three groups, each having the same set of samples and one of the three label sets. We ask the annotators to select all applicable labels from the label list, plus an extra ``None of the above / Others'' option. We also include a question of whether they feel restricted by the options and would use other words to describe the emotion(s). 

We took the set of 486 GoEmotions samples we used for the evaluation study (Section \ref{sec:sample_selection})
. We further removed samples that GPT-4 didn't output any candidates (N = 4) and defaulted the labels to ``neutral''.
We recruited 29 annotators for each group (i.e., Small, Large, and GPT-4 Pre-filtered) through crowdsourcing. We used the same crowdsourcing platform and setups as described in Section \ref{sec:human_preference_exp}. Each participant annotated 50 samples, and each sample got 3 annotations.

\subsubsection{Results}
\label{sec:pre_evaluation}
For the evaluation of the pre-filtering setup, we focus on three aspects: 1) cognitive load, indicated by subjective reports and time to completion; 2) label reliability, indicated by the agreement level among annotators; and 3) label coverage, i.e., the pre-filtered set should reasonably cover the labels human annotators selected from the Large set.

\begin{figure}[t]
  \centering
  \begin{subfigure}[b]{0.32\textwidth}
    \includegraphics[width=\textwidth]{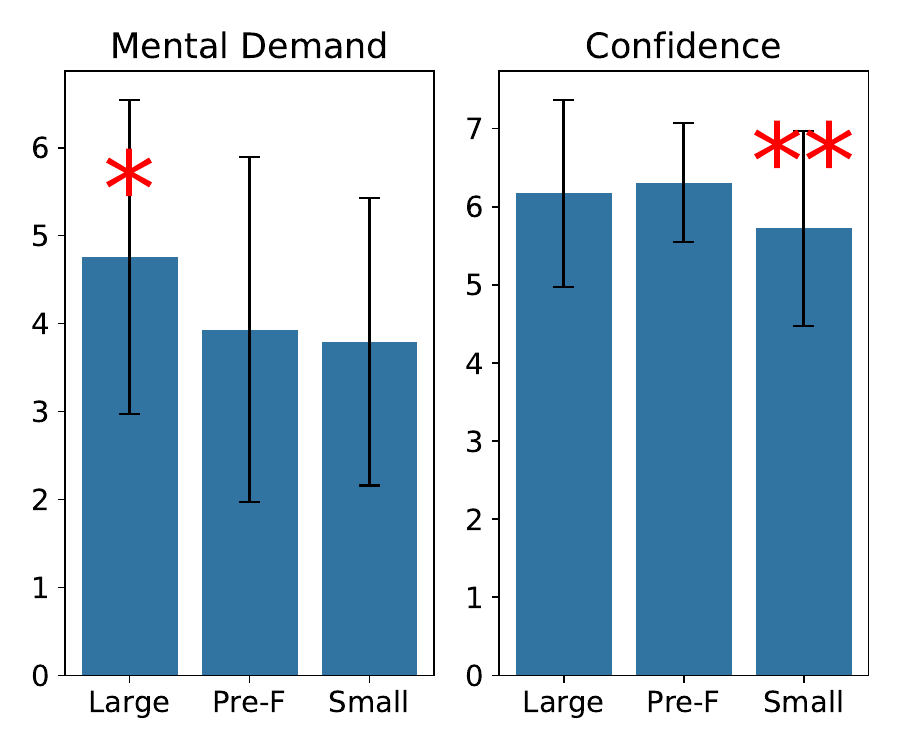}
    \caption{Subjective ratings}
    \label{fig:ratings}
  \end{subfigure}
  \hfill
  \begin{subfigure}[b]{0.16\textwidth}
    \includegraphics[width=\textwidth]{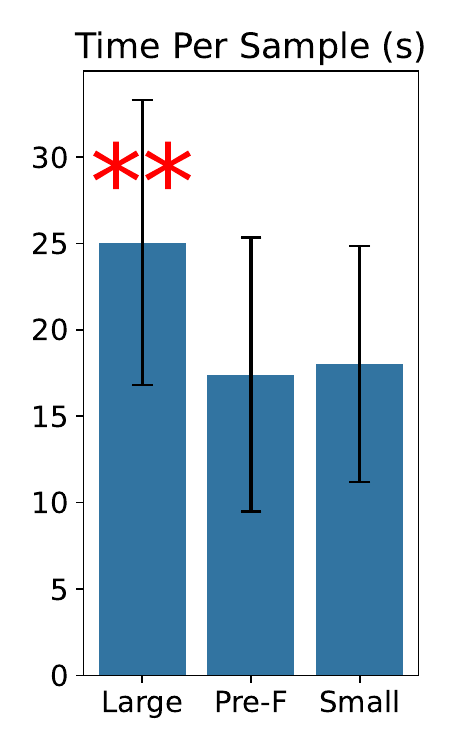}
    \caption{Time per sample}
    \label{fig:time}
  \end{subfigure}

  \caption{Comparison of the cognitive load on different label sets. The bars show the mean of the ratings, and error bars show the standard deviation. Significance tests are run between Pre-Filter and Large/Small sets. Dependent t-test is used in (a) because each annotator provided an overall rating for the whole session, and annotators were assigned different samples. Independent t-test is used in (b) because we measured the avg. time spent on each sample, and the sets share the same samples. \textcolor{red}{*}:p$<$0.1, \textcolor{red}{**}:p$<$0.05.}
\vspace{-10pt}
\end{figure}

\textbf{Cognitive Load and Annotators' Experience.} Following common approaches, we measured the cognitive load of the annotators in two aspects: perceived load\cite{hart2006nasa}, as a subjective measure, and time to completion\cite{wood2018comparison}, as an objective measure. We used the NASA Task Load Index\cite{hart1988development} as our cognitive load scale. We removed two questions that were not directly relevant in our task (physical demand and temporal demand), and we asked the participants to rate their feelings on four aspects on a 7-point scale at the end of their session: Mental Demand, Confidence, Effort, and Frustration. We found that a large label space significantly increased the mental demand of the annotators compared to the small set (see Figure \ref{fig:ratings}). However, a small label space did have a drawback: annotators reported feeling restricted by the options in 11.6\% of samples on the Large set, 13.9\% on Pre-filter, 33.5\% on Small. Consequently, significantly lower confidence was reported on the Small set (5.72$\pm$1.25, Large 6.17$\pm$1.20, Pre-filter 6.31$\pm$0.76). No significant differences were found in Effort and Frustration. We also compared the time the annotators spent on each sample. We excluded samples that took more than 60 seconds, as they were outliers in the time distribution and likely indicated a pause in the task. Annotators spent an average of 17.41 seconds on the Pre-filter set and 18.02 seconds on the Small set, while much longer (25.04 seconds) on the Large set (Figure \ref{fig:time}). 

\textbf{Agreement.} 
We use the Jaccard Index (JI)\cite{jaccard1912distribution} to measure the agreement between two annotators on each sample. JI is an agreement measure for multi-label classification tasks, defined by the size of the intersection of two label sets divided by their union. We calculated the average JI among pairs of annotators on each sample and the average across samples on each set. The Small set has the highest agreement of 0.29$\pm$0.34, slightly higher but not significantly different from the Pre-filter set (0.28$\pm$0.30, independent t-test p=0.36). The Large set has the lowest JI of 0.20$\pm$0.24, significantly lower than the Pre-filtered and Small sets (both p$<$0.05). We note that a smaller set tends to see higher agreement, so the JIs among sets are not directly comparable.
However, from a crowdsourcing perspective, reasonable agreement indicates reliability and is generally favorable\cite{nowak2010reliable}.

\textbf{Label Coverage.} 
Finally, we evaluate whether the filtering step retains potentially correct labels, i.e., the labels selected by humans in the Large set group. 
Following the approach of GoEmotions\cite{demszky2020goemotions}, we obtain aggregated labels from each set by using emotion classes that are selected by at least two annotators out of three. If no emotion class is selected for one sample, it is defaulted to ``neutral''. We first compared the chosen class labels with the Pre-filtered candidates, and found an average of 90.19\% of the labels selected in the Large set were included in the Pre-filter set, indicating a reasonably low false-negative rate. In addition, the labels annotators chosen from the Pre-filter set have an agreement of 0.30$\pm$0.32 JI with the Large set, which is comparable to (even slightly higher than) the within-set agreement levels
, and is much higher than the agreement between the Small and the Large sets (0.15$\pm$0.28, p$<$0.05). 

Together, the results show that the Pre-filtered set can match the advantages of both the Small and the Large sets: it maintains a low cognitive load and shorter time to completion (which often indicates lower cost) while allowing more descriptive accuracy and confidence of the annotators.

\subsection{Post-filtering, sample-level}
In Section \ref{sec:pre_evaluation}, we show some benefits of using GPT-4 for pre-annotation to collect new labels. In this section, we investigate a second approach: when a dataset with human-annotated labels is available, we propose to use GPT-4 as a quality checker to filter out potentially low-quality labels. Specifically, we compare the labels from human and GPT-4 annotation and drop the samples where the two sources totally disagree: i.e., they selected different labels for single-label classification datasets, or where they do not contain any overlapping labels for multi-label classification datasets. By applying this filtering step to GoEmotions, we obtained a much smaller Filtered set of 16,592 samples (out of 42,287).

\label{sec:model_training}
\begin{table}[t]
    \centering
    \begin{tabular}{cccccccc}
        \toprule
        \textbf{Model} & \textbf{\makecell{Test\\Label}} & \multicolumn{2}{c}{\textbf{Human}}  & \multicolumn{2}{c}{\textbf{Filter}} & \multicolumn{2}{c}{\textbf{Random\_F}} \\
        \cmidrule(lr){3-4} \cmidrule(lr){5-6} \cmidrule(lr){7-8} 
        & & \textbf{F1} & \textbf{UAR} & \textbf{F1} & \textbf{UAR} & \textbf{F1} & \textbf{UAR} \\
        \midrule
        BERT & H & \textbf{0.472} & 0.465  & 0.442 & \textbf{0.499} & 0.442 & 0.421\\
        & F & 0.578 & 0.530  & \textbf{0.620} & \textbf{0.590}  & 0.535 & 0.476 \\
        \midrule
        DBERT & H & 0.434 & 0.401  & \textbf{0.436} & \textbf{0.472} & 0.427 & 0.396\\
        & F & 0.526 & 0.462 & \textbf{0.588} & \textbf{0.551} & 0.504 & 0.441\\
        \bottomrule
    \end{tabular}
    \caption{BERT and DistilBERT model performance, finetuned and tested with different label sets. Test Label: H: Human, F: Filter. For training, the Human set has 42,287 samples and the Filter set has 16,592 samples. The ``Random\_F'' training set is a set randomly downsampled from the Human set to the size of the Filter set. Better performances are shown in bold, respectively for F1 and UAR. }
    \label{tab:model_performance}
\end{table}

\subsubsection{Evaluation Setup}
Since the post-filtering step removes samples where GPT-4 and human annotations disagree, it is expected to remove samples with mistakes in annotations, resulting in higher-quality labels. While this generally benefits model training, this filtering step also decreases the number of samples and potentially the diversity or ambiguity in the samples. Therefore, a key question is whether this trade-off eventually enhances or hurts model training outcomes.
To evaluate this, we train smaller models with either the whole labeled GoEmotions training set, or the smaller Filtered set. We measure the performance on its test set as an indicator of the usefulness of the labels. We report performance on both the whole test set and a filtered test set.

\textbf{Base model selection.} We choose two models from the BERT family for our training experiments: BERT \cite{devlin2018bert} and DistilBERT\cite{sanh2019distilbert}. BERT is one of the earliest transformer-based LLMs that gained broad attention, and it has been used as a baseline for many NLU tasks\cite{koroteev2021bert}, including in the GoEmotions paper\cite{demszky2020goemotions}. With 110 million parameters, BERT is significantly smaller than leading LLMs like GPT-4, making it practical for use on most modern GPUs. DistilBERT is a distilled version of the BERT model with a 40\% reduction in the number of parameters while delivering comparable performance in multiple NLU tasks. We compare the models trained on the original human labeled set versus the Filtered set where samples that GPT-4 totally disagree with are dropped. We finetune the models for 30 epochs with a learning rate of 1e-5, and we select the best model measured by performance on the validation set. We use our annotated ground-truth set as the primary test set for evaluation, while we also report the performance on the test split of the human and post-filtered set. 

\subsubsection{Results}
Results are presented in Table \ref{tab:model_performance}. The filtered set, despite comprising less than 40\% of the samples in the full set, consistently led to better model performance across models (both BERT and DistilBERT) and both the full and filtered test sets (with one exception of the F1 score when tested in-domain on the human-labeled set). To isolate the effect of training sample size reduction, we included a training set that was randomly sampled from the Human set and matched the size of the Filter set (16,592). As expected, this smaller set led to a performance drop, with all metrics lower compared to the full Human set. This further highlights the effectiveness of our post-filtering approach, which achieved better performance with much fewer samples. Together, these results show the potential of GPT-4 to flag possibly low-quality samples, thus improving model performance as well as training efficiency.


\section{Discussion}
\label{sec:discussion}
Our work examines many design choices involved in emotion annotation and investigates how LLMs, specifically GPT-4, perform in this context and where they may offer new possibilities. In the first part of our study (Section \ref{sec:evaluation_study}), we evaluated GPT-4’s ability to classify emotions across three datasets with varying domains and label space complexity. We found that GPT-4 predictions generally align with human-annotated labels. In addition, a human evaluation study revealed preferences for GPT-4 labels over original human annotations, highlighting the value of human-centered evaluations and raising questions about whether human labels should be used as the sole ground-truth for evaluating LLMs. We also compared GPT-4 and human labels across different label spaces. Results suggest that larger label spaces allow nuanced emotion descriptions, which are perceived as more accurate by human evaluators, while smaller spaces are less cognitively demanding and can potentially lead to fewer human mistakes. 

In the second part, we further explored ways to integrate GPT-4 into annotation processes, focusing on the GoEmotions dataset. We found that GPT-4 can serve as a pre-annotation label filter to dynamically exclude highly unlikely labels before presenting them as options to human annotators. Our human annotation study showed that, compared to traditional methods, GPT-4 could effectively reduce more than 70\% of options while preserving more than 90\% of human-selected labels. This approach leverages the expressivity of larger label spaces and the reduced cognitive load and higher annotator agreement associated with smaller label spaces. What's more, on annotated datasets, GPT-4 can act as a post-annotation sample filter to flag potentially low-quality labels. Models trained on the filtered dataset, although much smaller in training data size, achieved better performance than the original full set with human annotations.

\section{Limitations and Future Work}
Through our studies, we encountered several challenging factors that can influence the collection and evaluation of emotion labels. First, our evaluation study (Section \ref{sec:evaluation_study}) reveals that using human labels as ground truth is not always reliable. However, we do not have a good alternative evaluation approach that can account for the inherent subjectivity in emotion. This limitation influenced our quantitative model evaluation in Section \ref{sec:model_training}, where we relied on a human-annotated set created through multi-annotator discussions. Although we believe that this annotation method greatly reduces the possibility of mistakes, we also note the ambiguity of many samples, and thus reasonable alternative interpretations are possible. Developing new evaluation metrics that address these complexities would be a valuable, though challenging, direction for future research.

We also limited our discussion to classification tasks. Our previous work conducted preliminary experiments that included dimensional labels with the Emobank dataset\cite{niu2024text} and found that the scale of GPT-4 output did not quite align with human labels, raising more questions for evaluation. LLMs are generally good at language-based interactions, and language anchors are helpful for them to understand dimension scales\cite{gong2023joint}. As dimensional label spaces gain more popularity\cite{buechel2016emotion}, future research could explore ways to better leverage LLMs in dimensional emotion annotation.

Additionally, our pre- and post-filtering methods (Section \ref{sec:integration_ways}) serve as preliminary demonstrations of the feasibility and potential of GPT-4-assisted annotation rather than as definitive solutions. Future work could incorporate more refined approaches to further improve performance. While our focus is not on comparing different perspectives or methods in human annotation processes (e.g., self-reported vs. third-person annotations, or crowdsourcing vs. in-house annotations), previous studies have examined these aspects\cite{biersack2005tracing, busso2008expression, kim2021label}. Finally, prompting techniques are not the focus of this paper, but we acknowledge the sensitivity of the models and the importance of good prompts. We direct interested readers to related studies\cite{loya2023exploring, Liu2023-yz, binz2023using, wei2022chain}, and we encourage new explorations with our prompts and code publicly available~\footnote{https://github.com/chailab-umich/GPT-4-Emotion-Annotation}.

\section{Conclusion}
In this work, we conduct a comprehensive evaluation of GPT-4's emotion classification performance and its potential to aid annotation processes. We present encouraging results along with discussions on the complexities and challenges associated with various design choices in emotion annotation studies. Our findings underscore the importance of carefully rethinking these choices with LLMs' capability in mind. We highlight the need for evaluation metrics that better align with human perspectives and the strong promise of using LLMs as tools to aid annotation efforts.

\section{Acknowledgments}
This material is based in part upon work supported by the National Science Foundation (NSF IIS-RI 2230172) and National Institutes of Health (NIH R01MH130411).

\ifCLASSOPTIONcaptionsoff
  \newpage
\fi



\bibliographystyle{IEEEtran}
\bibliography{main}
\end{document}